\documentclass[letterpaper, 10 pt, conference]{ieeeconf}  %

\IEEEoverridecommandlockouts
\usepackage{graphicx}
\usepackage{tabularx}
\usepackage{amsmath} 
\usepackage{amssymb}  
\usepackage{caption}
\usepackage{subcaption}
\usepackage{gensymb}
\usepackage{svg}
\usepackage[hyphens]{url}
\usepackage{hyperref}
\usepackage[noadjust]{cite}

\usepackage{color, soul} 
\renewcommand\hl[1]{#1} 

\usepackage{xcolor}

\title{\LARGE \bf Self-supervised Transparent Liquid Segmentation for Robotic Pouring}

\author{Gautham Narasimhan$^{1}$, Kai Zhang$^{2}$, Ben Eisner$^{1}$, Xingyu Lin$^{1}$, David Held$^{1}$

\thanks{$^{1}$ Gautham Narasimhan, Ben Eisner, Xingyu Lin, and David Held are affiliated with the Robotics Institute at Carnegie Mellon University, Pittsburgh, PA. {gauthamnarayn@gmail.com,} \{baeisner, xlin3, dheld\}@andrew.cmu.edu}%

\thanks{$^{2}$ Kai Zhang is affiliated with the University of Notre Dame.}

}

\begin{document}

\maketitle
\thispagestyle{empty}
\pagestyle{empty}

\begin{abstract}

Liquid state estimation is important for robotics tasks such as pouring; however, estimating the state of transparent liquids is a challenging problem. We propose a novel segmentation pipeline that can segment transparent liquids such as water
from a static, RGB image without requiring any manual annotations or heating of the liquid for training. Instead, we use a generative model that is capable of translating images of colored liquids into synthetically generated transparent liquid images, trained only on an unpaired dataset of colored and transparent liquid images. Segmentation labels of colored liquids are obtained automatically using background subtraction.  
Our experiments show that we are able to accurately predict a segmentation mask for transparent liquids without requiring any manual annotations. We demonstrate the utility of transparent liquid segmentation in a robotic pouring task that controls pouring by perceiving the liquid height in a transparent cup. Accompanying video and supplementary materials can be found at \url{https://sites.google.com/view/transparentliquidpouring}.

\end{abstract}

\section{INTRODUCTION}
\label{sec:introduction}


Robots that could pour liquids would enable us to automate tasks such as cooking, pouring medicines into vials in pharmacies, or watering our plants. 
However, transparent liquids are difficult to perceive in images; the only visual signals a perfectly transparent liquid can provide are the refraction of light passing through the liquid. Obtaining depth measurements for liquids is similarly difficult since the liquid will refract the projected infrared light.

\begin{figure}
    \centering
    \includegraphics[width=0.8\columnwidth]{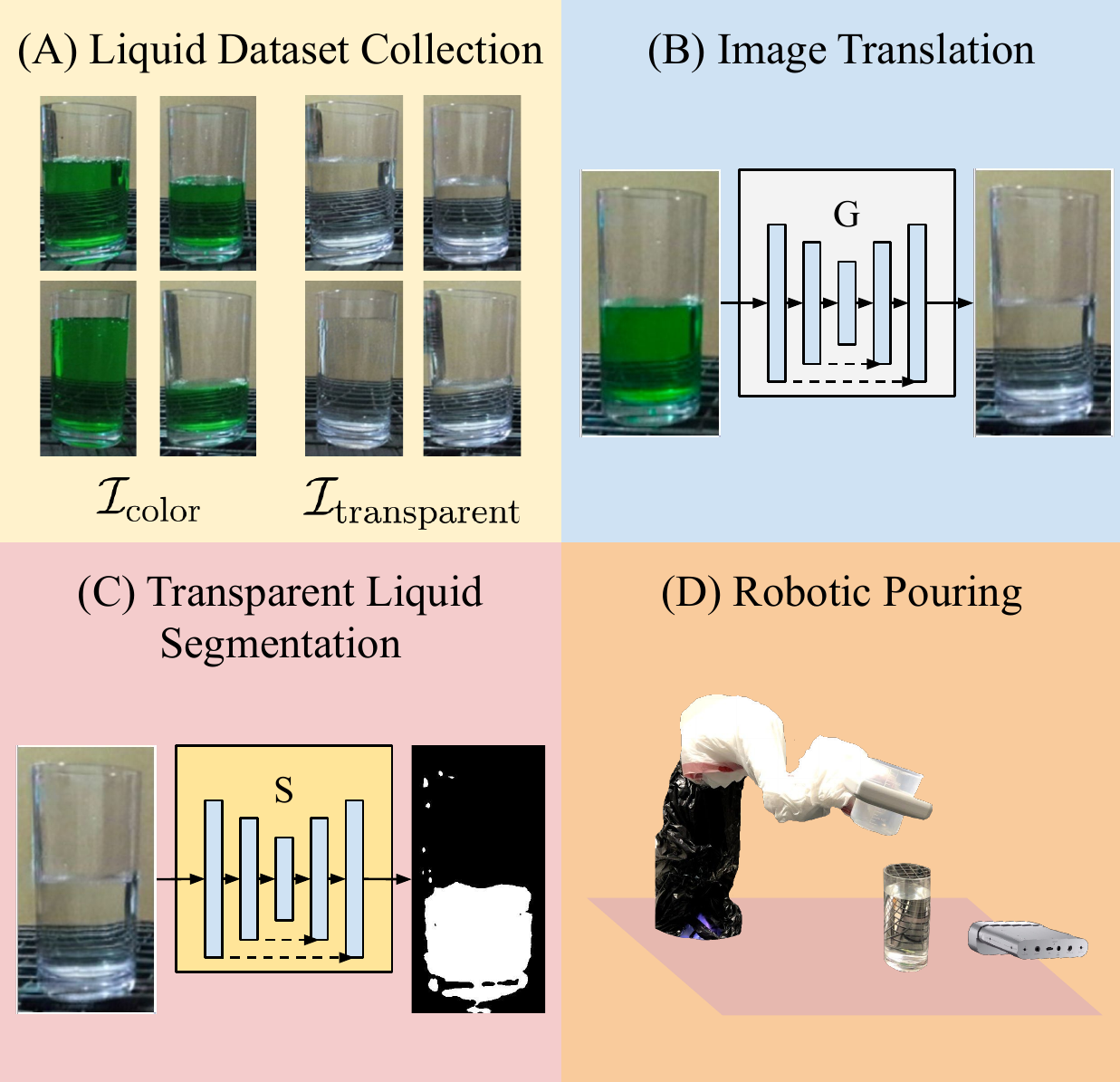}
    \caption{We introduce a simple method for learning to segment transparent liquids in transparent containers.  Our approach consists of four steps: (A) collect two datasets (unpaired) of colored and transparent liquid in containers; (B) create synthetic segmentation labels for transparent liquids using image translation (C) train a transparent liquid segmentation model using the generated labels; (D) closed-loop robotic pouring using our transparent liquid segmentation. }
    \label{fig:teaser}
    \vspace{-2em}
\end{figure}





Previous works have explored robotic pouring in various settings~\cite{liquid_level_detection,precision_pouring_chenyu,precise_dispensing,autonomous_pouring_precision_containers,matl2020inferring}, 
but all require significant compromises in the environment or method for data collection. Several methods for transparent liquid segmentation require heating up the liquid during training to obtain ground-truth labels when viewed by a thermal camera~\cite{schenck_percieve_and_reason,schenck_visual_closed_loop_control,schenck_reasoning_closed_loop_sim}; however, heating up the liquid for training is a tedious process that limits how much training data can be easily collected.  Other approaches require observing the liquid from multiple viewpoints~\cite{liquid_level_detection}, checkerboard backgrounds~\cite{autonomous_pouring_precision_containers}, weight measurements~\cite{autonomous_pouring_precision_containers}, or liquid motion~\cite{atkeson_stereo_vision,liquid_pouring_sequences_audio_visual,liquid_height_audio_only,accurate_pouring_rgbd}; these requirements on the environment restrict the applicability of these methods. 

In this work, we propose a method for perceiving transparent liquid (such as water) inside transparent containers.  Our method eases restrictions on the operational domain as compared to previous methods. 
Specifically, our method operates on individual images (we do not require liquid motion or multiple frames), and requires no manual annotations or heating of liquids during training. Instead, our method uses a generative model that learns to translate images of colored liquid into synthetic images of transparent liquid, which can be used to train a transparent liquid segmentation model.

Importantly, we train this translation model on an \textit{unpaired} image dataset of transparent and colored liquids - that is, our method does not need labeled correspondence between colored images and transparent images to learn the colored-to-transparent translation model. This enables automatic and highly-efficient dataset collection. Because it is easy to obtain segmentation labels for colored liquids (e.g. using background subtraction), our method allows us to directly use segmentation labels from an image of a colored liquid as a ground-truth segmentation label for that same image once it has been translated into a transparent liquid.

To demonstrate the utility of this dataset translation method in a real-world system, we construct a robotic pouring system that utilizes a transparent liquid segmentation model to accomplish a pouring task. We train this segmentation model on a small dataset of synthetic transparent liquid images generated by our translation model in our robot's workspace. Finally,
we conduct several dataset augmentation experiments to demonstrate the potential of our method to train transparent liquid segmentation models that generalize to diverse scenes.

\section{Related Work}
\label{sec:related_work}

\subsubsection{Transparent object perception} Perceiving transparent objects is particularly challenging because transparent objects can 
refract, reflect, and absorb light. 
Some previous works focus on perceiving transparent containers; methods have been developed for transparent object segmentation~\cite{translab,transparent_casually_captured_videos}, depth estimation~\cite{cleargrasp,rgbd_implicit_function}, keypoint estimation~\cite{liu2020keypose}, and transparent object matting~\cite{tomnet}.  Other methods for segmenting transparent objects use light field cameras~\cite{transcut,transcut2} or light polarization~\cite{deep_polarization}. On the manipulation side, other recent works have been developed to directly grasp transparent objects without first estimating their 3D shape~\cite{tweng2020multi}. Our approach builds on~\cite{translab} for transparent container segmentation; 
 however, our focus is on segmenting the transparent liquid inside the container. Unlike previous work that uses manual annotations for training~\cite{translab}, our work does not rely on any manual annotations.

\subsubsection{Transparent liquid perception} 
Perception of liquid is more challenging than perception of object due to the lack of a fixed shape or geometry. While perception of colored liquid can sometimes be done using background subtraction~\cite{precise_dispensing}, it does not work for transparent liquid.
One approach to transparent liquid perception is to use heated liquid observed by a thermal camera to obtain ground-truth labels for liquid~\cite{schenck_percieve_and_reason,schenck_visual_closed_loop_control,schenck_reasoning_closed_loop_sim}. However, the requirement to heat the liquid before recording the ground-truth is a tedious process; our method does not require heated liquid.
To segment liquid while it is being poured, one can use optical flow~\cite{atkeson_stereo_vision} or audio signals~\cite{liquid_pouring_sequences_audio_visual,liquid_height_audio_only}. Our method can segment static liquid, which is important for liquid state estimation before initiating a pouring task. Some methods reason about the refraction of the infrared light emitted by a depth sensor~\cite{liquid_level_detection,accurate_pouring_rgbd}, multiple noisy readings from different viewpoints~\cite{liquid_level_detection}, or from different time points during pouring~\cite{accurate_pouring_rgbd}, integrated probabilistically.  In contrast, our method can segment the liquid from just a single RGB image. Another approach is to use a depth sensor to estimate the height of the liquid surface~\cite{precision_pouring_chenyu}; however, such depth readings are inaccurate for transparent liquids. A different strategy is to pour liquid in front of a checkerboard background
or to use weight readings from a scale~\cite{autonomous_pouring_precision_containers}. Our method does not require a checkerboard background or a scale. Finally, some approaches forgo a separate module for transparent liquid perception and learn an end-to-end policy for pouring transparent liquid~\cite{lin2020softgym}. However, so far such approaches have only been shown to work in simulation due to the sample complexity of learning a sensorimotor policy.

\section{METHOD}
\label{sec:method}
We describe our method for transparent liquid segmentation when liquids are placed within transparent containers (see Figure~\ref{fig:vision_system} for an overview). First, we collect a dataset of colored liquid and another (unpaired) dataset of transparent liquid.  We then use an image translation method to learn to translate an image of colored liquid into a synthetically generated image of transparent liquid that is identical to the input image, except that the liquid is now transparent.  Next, we use background subtraction to find the colored liquid pixels in the colored liquid dataset.  We treat the colored liquid segmentation as a ground-truth label for the synthetically generated transparent liquid. We then train a network to segment transparent liquid, using paired samples of the synthetically generated transparent liquid and colored liquid ground-truth labels. 


\subsection{Learning to translate colored liquid to transparent liquid}
\label{sec:dataset_translation}

\begin{figure*}[t]
    \centering
    \begin{subfigure}[b]{0.49\textwidth}
        \centering
        \includegraphics[width=0.8\columnwidth]{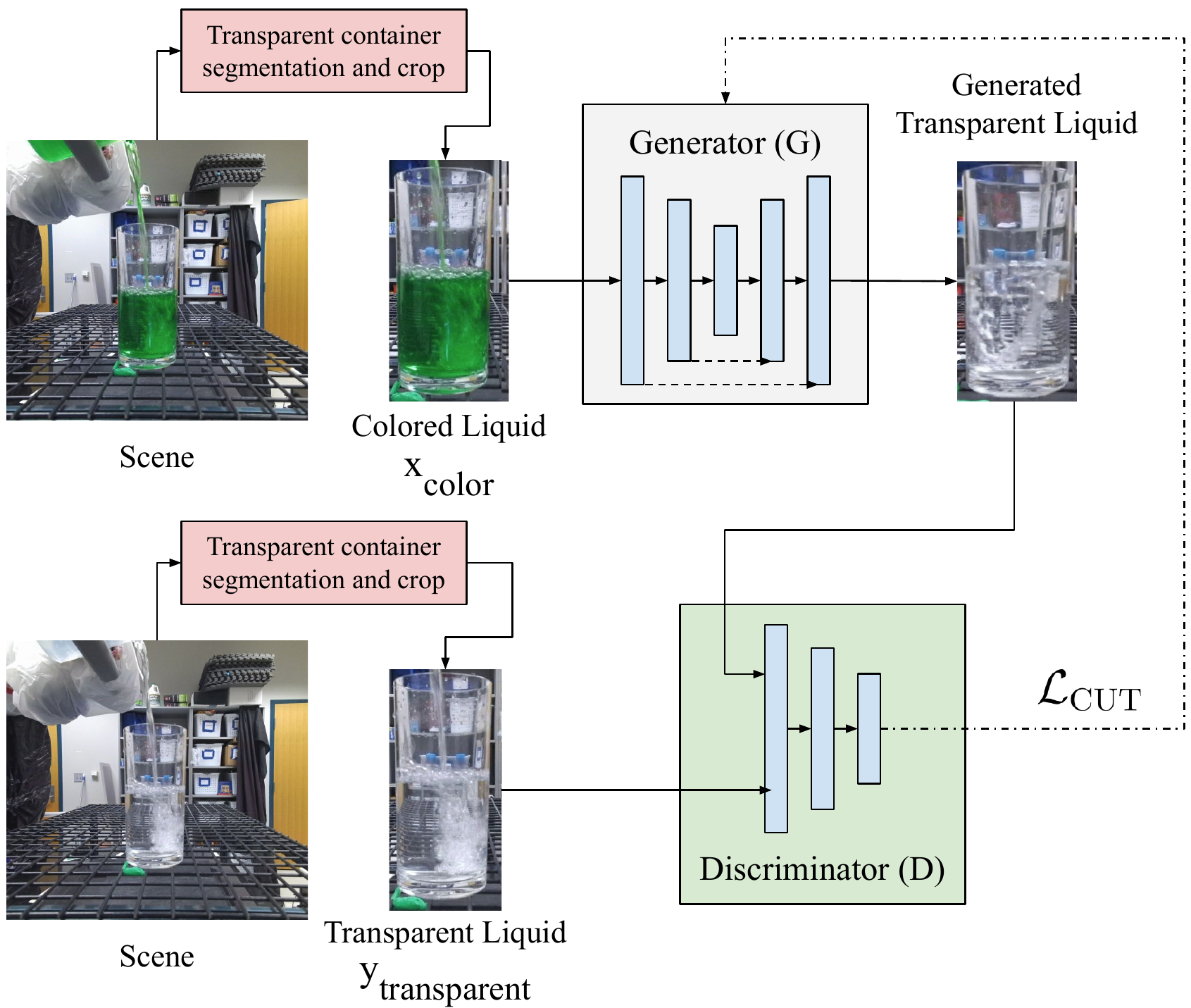}
        \caption{Image Translation}
        \label{fig:cut}
    \end{subfigure}
    \hfill
    \vrule
    \hfill
    \begin{subfigure}[b]{0.49\textwidth}
        \centering
        \includegraphics[width=0.8\columnwidth]{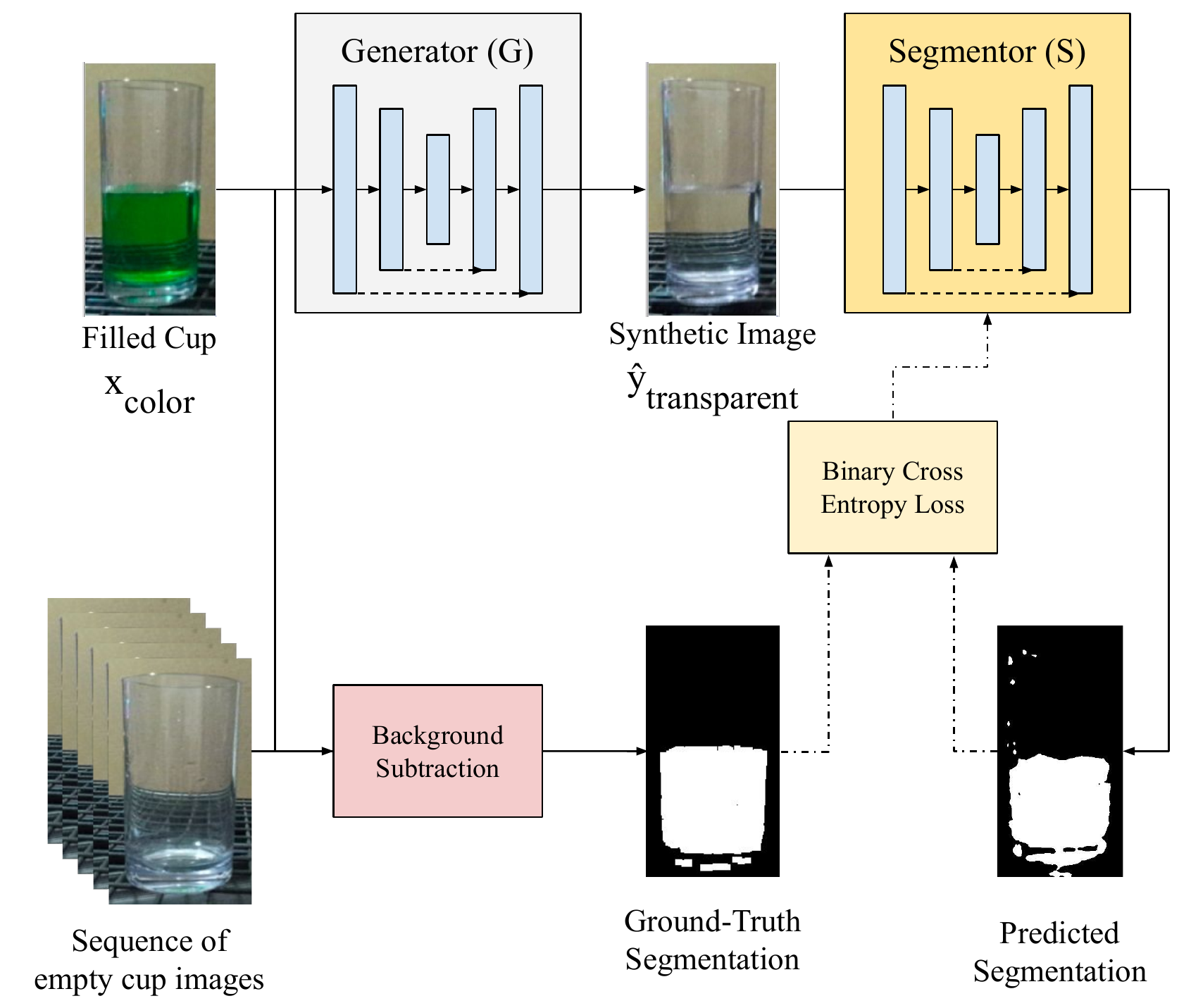}
        \caption{Transparent Liquid Segmentation}
        \label{fig:segmentation}
    \end{subfigure}
    \caption{Our vision pipeline for training a segmentation network that can be used to segment transparent liquids. In (a), we use the losses described in Section \ref{sec:dataset_translation} to train a generator $G$ which transforms images of colored liquids in $\mathcal{D}_{\mathrm{color}}$ to look like images in $\mathcal{D}_{\mathrm{transparent}}$. In (b), we use this generator $G$ to translate images in $\mathcal{D}_{\mathrm{color}}$ into transparent images, and we generate pseudo-ground truth segmentation masks $M_i$ from the colored liquid via  background subtraction. Finally, we train a transparent liquid segmentation model $S$ on this synthetic dataset using a standard binary cross-entropy loss.} 
    \label{fig:vision_system}
\end{figure*}

To train a model for transparent liquid segmentation, most standard supervised segmentation learning methods require a dataset of labeled images of transparent liquids.  However, labeling such a dataset is tedious.  Instead, we make use of an image translation method to synthetically generate the desired labels.

We collect one dataset of colored liquids in transparent containers $\mathcal{D}_{\mathrm{color}}$ and a second dataset of transparent liquids in transparent containers $\mathcal{D}_{\mathrm{transparent}}$. Importantly, these datasets are \textit{unpaired}, meaning that there is no explicit alignment or correspondence between examples across the datasets, and thus each can be collected totally independently. Given these two datasets, we learn an image translation model from colored to transparent liquids. To do so, we use Contrastive Unpaired Translation (CUT) ~\cite{cut}, which we train to convert an image of a colored liquid into an image of a transparent liquid.


We briefly describe CUT and how we adapt it for our method. The backbone of CUT is a generator that translates an image of the source domain into an image of the target domain. To encourage this translation, three loss terms are used. First, a standard adversarial GAN loss is used to encourage the generator to output images that are visually similar to those in the target domain:
\begin{equation}
    \mathcal{L}_{\mathrm{GAN}} =  \mathbb{E}_{y \sim Y} \left[ \log D(\boldsymbol{y}) \right] + \mathbb{E}_{x \sim X} \left[\log (1-D(G(\boldsymbol{x})))\right]
    \label{eq:gan_loss}
\end{equation}
where $G, D,$ denote the generator and the discriminator respectively, and $X$, $Y$ denote the source domain and the target domain, respectively. The generator $G$ is divided into an encoder $G_\textrm{enc}$ and a decoder $G_\textrm{dec}$, such that the output is $\hat{y} = G(x) = G_\textrm{dec}(G_\textrm{enc}(x))$, for an image $x$ from the source domain.

Additionally, CUT uses a patch-wise contrastive loss \cite{oord2019representation} to encourage corresponding patches between the input and output images to be similar to each other in feature space. Specifically, given an image $x$ from the source domain $X$, the image is translated into an image $y$ of the target domain $Y$.  The patch-wise contrastive loss, $\mathcal{L}_{\text{PatchNCE}}(G, H, X)$, maximizes the mutual information between $H(G_\textrm{enc}(x))$ and $H(G_\textrm{enc}(y))$, where $H$ is a small multi-layer perceptron~(MLP). The generator is also trained with an identity loss $\mathcal{L}_{\text{PatchNCE}}(G, H, Y)$ to help regularize the encoder and minimize unnecessary modifications to a source image. The combined loss is:
\begin{equation}
\begin{split}
    \mathcal{L}_{\mathrm{CUT}} = \mathcal{L}_{\mathrm{GAN}} & + \lambda_{X} \mathcal{L}_{\text{PatchNCE}}(G, H, X) \\
    & + \lambda_{Y} \mathcal{L}_{\text{PatchNCE}}(G, H, Y)
\end{split}
\end{equation}

We apply CUT directly to our two datasets of raw images, where our source domain $X = \mathcal{D}_{\mathrm{color}}$ and target domain $Y = \mathcal{D}_{\mathrm{transparent}}$. Thus, we use CUT to convert an image of colored liquid $x_{\mathrm{color}} \in \mathcal{D}_{\mathrm{color}}$ to a synthetically generated image of a transparent liquid $\hat{y}_{\mathrm{transparent}} = G(x_{\mathrm{color}})$.

Importantly, the patch-wise contrastive loss encourages the object parts in the image $x_{\mathrm{color}}$ to be in the same location as the parts in the translated image $\hat{y}_{\mathrm{transparent}}$.  For example, in Figure~\ref{fig:cut}, the cup, liquid, surface, and even shadows are in the same locations between the image of colored liquid $x_{\mathrm{color}}$ that is input to the generator $G$, compared to the synthetic image of transparent liquid $\hat{y}_{\mathrm{transparent}}$ that is output by the generator.  The primary difference between these images is that the colored liquid has changed to become transparent; the liquid is in the same location as in the input.

This property (that the liquid is in the same location in the generator input $x_{\mathrm{color}}$ as in the output $\hat{y}_{\mathrm{transparent}}$) is crucial to the success of our proposed segmentation method. 
If we assume that the only property that has changed as a result of the translation is the liquid color, then we can directly use the segmentation masks from the colored liquid $x_{\mathrm{color}}$ as pseudo-ground truth for the generated transparent liquid $\hat{y}_{\mathrm{transparent}}$. Because it is simple to segment an image of colored liquid using color thresholding or background subtraction (see supplementary materials for details), we can then easily generate segmentation labels for the synthetic transparent images $\hat{y}_{\mathrm{transparent}}$ without requiring human annotations. Formally, given an image of colored liquid $x_{\mathrm{color}}^{(i)}$ with corresponding segmentation mask $M^{(i)}$, we generate a synthetic image of transparent liquid $\hat{y}_{\mathrm{transparent}}^{(i)} = G(x_{\mathrm{color}}^{(i)})$ to which we associate the colored image's segmentation mask $M_i$ as a pseudo-ground truth segmentation label. This creates a synthetic labeled dataset: $\{\hat{y}_{\mathrm{transparent}}^{(i)}, M^{(i)}\}$ which we use to train our transparent liquid segmentation model.


\subsection{Learning transparent liquid segmentation}
\label{sec:segmentation}

We can use the aforementioned synthetic labeled dataset to train a transparent liquid segmentation model $S$
using the Binary Cross Entropy loss between the predicted liquid segmentation mask and the pseudo-ground truth $M_i$ (described above). 
Architectural, hyper-parameter, and implementation details are described in the supplementary materials. 

\subsection{Robot liquid pouring}
\label{sec:method_pouring}

\begin{figure}[ht]
    \centering
    \includegraphics[width=0.8\columnwidth]{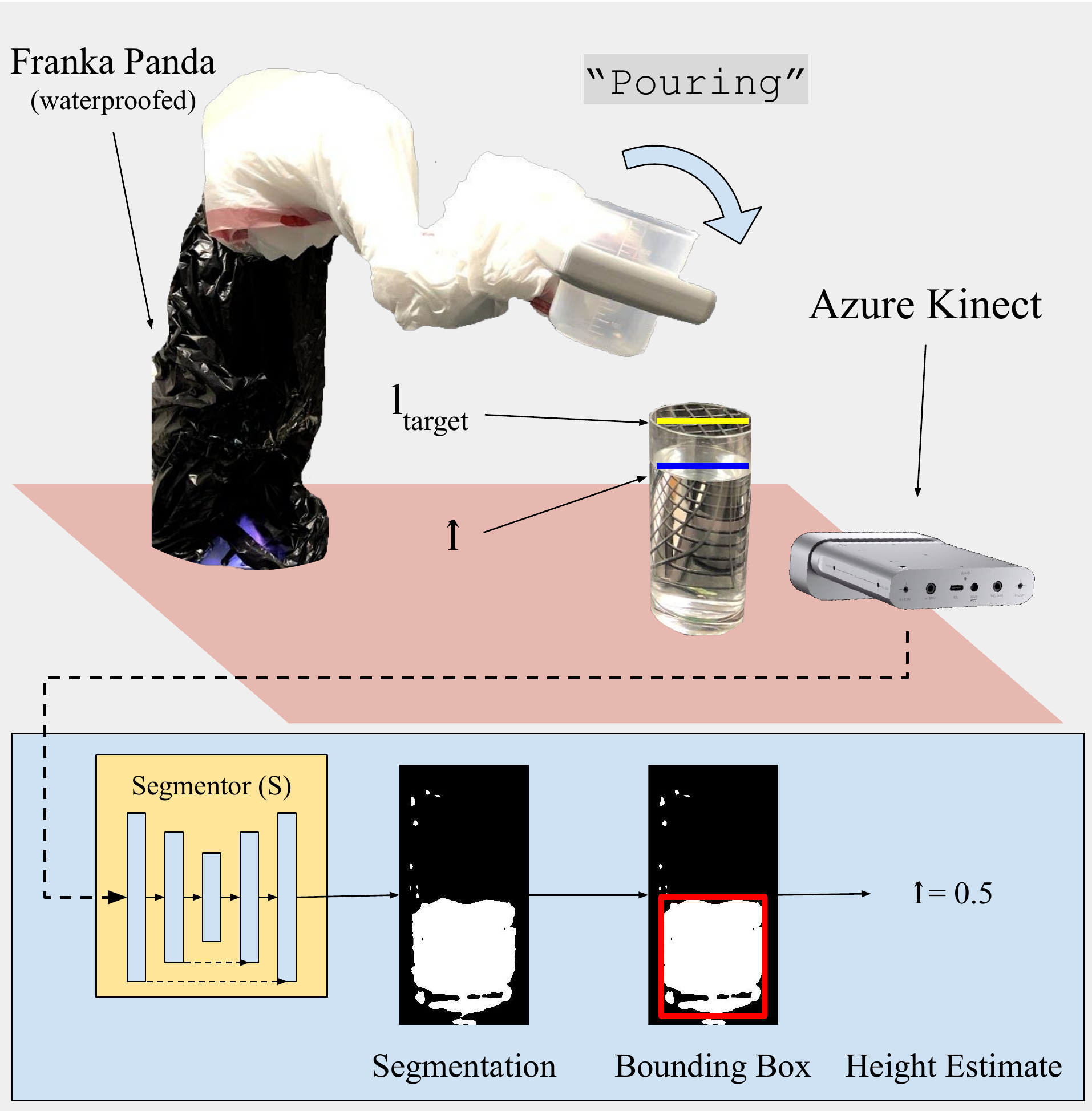}
    \caption{Our robotic pouring system. For each RGB image captured, we use our learned segmentation model $S$ to output a segmentation of the liquid in the cup, and we detect the bounding box of the cup using TransLab \cite{translab}. We compute a liquid level estimate $\hat{l}$, and we pour at a fixed angle until the perceived liquid level is within a small threshold.}
    \label{fig:robot_pouring_system}
    \vspace{-2em}
\end{figure}

In this section, we describe the robotic pouring system we designed to demonstrate the utility of our transparent liquid segmentation model in pouring tasks. While other works have explored sophisticated and flexible robotic pouring methods, we emphasize that our robotic system design is a simple testbed for our perception method. Our system consists of two stages: a visual postprocessing stage that converts a liquid segmentation  into an estimate of a fill level in a container $\hat{l}$ and a pouring controller which drives the system to reach a target fill level $l_{\mathrm{target}}$ (see Figure \ref{fig:robot_pouring_system}).

\subsubsection{Task Design} For our robotic pouring experiments, we use a Franka-Emika Panda 7-DOF robotic arm, with a custom-built end-effector that rigidly secures a source container for pouring \cite{bronstein_kang_kolligs_liao_alini_2021}.
We place a Microsoft Azure Kinect RGB camera directly in front of the robot's workspace. We place the target cup directly in front of the camera at a known location.  We then drive the end-effector to a known location directly above the target cup. We ensure that the source cup in the robot's end-effector is not included in the image recorded by the Kinect camera.  We fill the source cup with liquid, which will be poured into the target cup.


\subsubsection{Visual postprocessing} 
To minimize any potential errors caused by the segmentation model, we first use an off-the-shelf method (TransLab~\cite{translab}) to detect the location of the transparent cup in the scene.  Next, we crop the image region around the detected cup location. Then, we use our model to segment the transparent liquid pixels from within this crop. Finally, we perform a filtering step \hl{(see supplementary materials for details)} that removes noise as well as small water particles in motion (i.e. during pouring). The remaining segmented points are from liquid inside the target cup.

\subsubsection{State estimation from segmentation} 

It is difficult to design a control system for pouring that operates directly on a current and target segmentation mask.  Instead, we define a target fill height $h_\mathrm{target}$ from the camera perspective; we then define the process variable as the fraction of the cup that is filled: $l_{\mathrm{target}} = \frac{h}{h_{\mathrm{cup}}}$, where $h$ is the liquid height and $h_{\mathrm{cup}}$ is the cup height (as seen from the camera perspective).  This task was inspired by the task of a robot waiter, which must refill everyone's cups to a certain fraction of the cup height. 
Further, keeping the process variable grounded in 2D visual features avoids complexities of estimating the 3D geometry of the scene.

To estimate the current fill level $\hat{h}$,
we compute a vertically-aligned bounding box of the segmented pixels.  We then use the upper edge of the box as the estimated height $\hat{h}$ of the liquid. As mentioned previously, we use previous work to segment the transparent cup in the scene~\cite{translab}; we use a similar approach as above to fit a bounding box and find the height of the cup $h_{\mathrm{cup}}$. From these computations, we can estimate the process variable  $\hat{l} = \frac{\hat{h}}{h_{\mathrm{cup}}}$.


\subsubsection{Control} With the robot arm positioned directly above the cup, we restrict the control space to be one of two states: $\left\{ \texttt{NotPouring}, \texttt{Pouring} \right\}$. The \texttt{NotPouring} state corresponds to the starting postion, where the source cup is completely vertical (upright). The \texttt{Pouring} state corresponds to a 60\degree~rotation about the central axis of the cup (parallel to the table). When the control signal changes between the two states, the end-effector rotates at a constant angular velocity until the desired control state is reached. 
The system initially commands \texttt{Pouring} until the estimated height is within a margin $\epsilon$ of the target; afterwards, the system commands \texttt{NotPouring}.

Because the vision system is a simple neural network with simple post-processing, it has very low latency.
Therefore, we can operate our control loop at roughly 10Hz. Because of the responsive system, bang-bang control of the pouring into the target cup is effective. Finally, to compensate for errors that occur in the perception system when the target cup is empty, we always begin the control with roughly 1s of initial pouring.

\section{RESULTS}

Our evaluation consists of two parts. First, we  demonstrate how our method for synthetic dataset generation can be used to quickly create a perception module that can be used in a robotic pouring setting. Second, we seek to characterize the ability of our method to handle diverse transparent liquid scenarios such as variations in cup shape and image background. 

\subsection{Full-System Evaluation}

We conduct an analysis of our robotic pouring system, considering the performance of each module independently as well as an end-to-end  analysis of the full system. The models we train in this setting are specific to our robot environment, though in Section~\ref{sec: generalization} we provide an analysis of how our method can generalize to new settings. 

\subsubsection{Dataset Description} To train our method for use in our robot's workspace, \hl{we collected 15 distinct pouring videos each of green-colored water and clear water. This resulted in datasets with 2231 and 2237 RGB frames, respectively. Our method does not require the images to be aligned or paired between datasets. We also collected a test set of 133 images across 4 different videos of pouring transparent liquids in the same scene.}
We manually annotate the location of the transparent liquids for the test set using Labelme~\cite{labelme};  however, we do not use such annotations for training.

\subsubsection{Visual Translation}

\begin{figure}[ht]
    \centering
    \begin{subfigure}[b]{0.15\columnwidth}
        \centering
        \includegraphics[width=\linewidth]{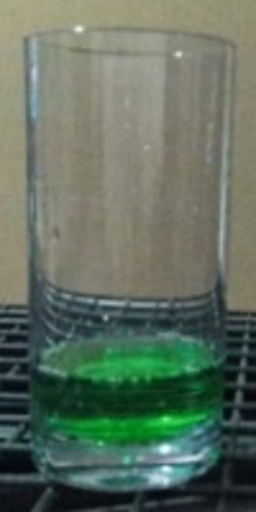}
    \end{subfigure}
    \begin{subfigure}[b]{0.15\columnwidth}
        \centering
        \includegraphics[width=\linewidth]{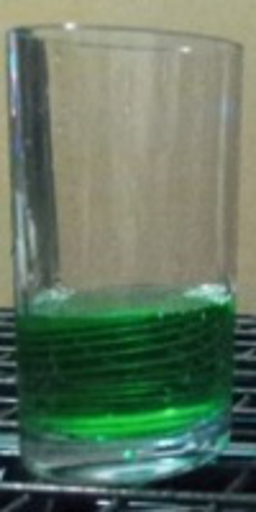}
    \end{subfigure}
    \begin{subfigure}[b]{0.15\columnwidth}
        \centering
        \includegraphics[width=\linewidth]{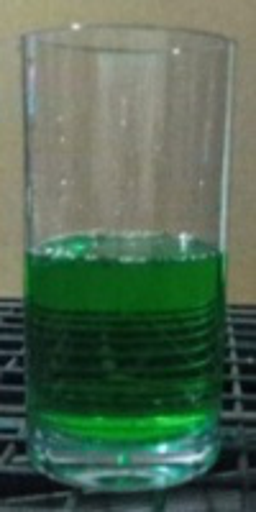}
    \end{subfigure}
    \begin{subfigure}[b]{0.15\columnwidth}
        \centering
        \includegraphics[width=\linewidth]{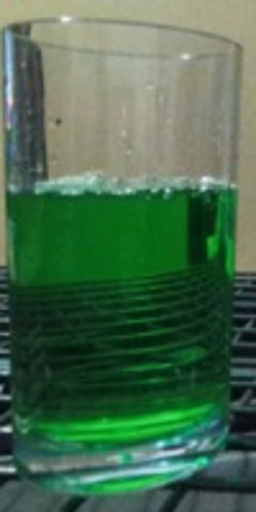}
    \end{subfigure}
    \begin{subfigure}[b]{0.15\columnwidth}
        \centering
        \includegraphics[width=\linewidth]{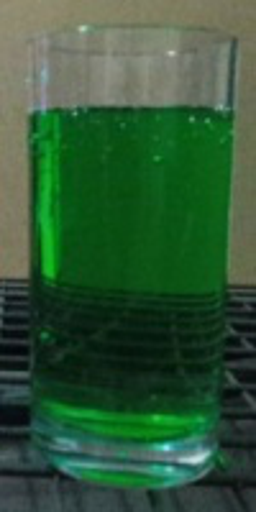}
    \end{subfigure}
    \begin{subfigure}[b]{0.15\columnwidth}
        \centering
        \includegraphics[width=\linewidth]{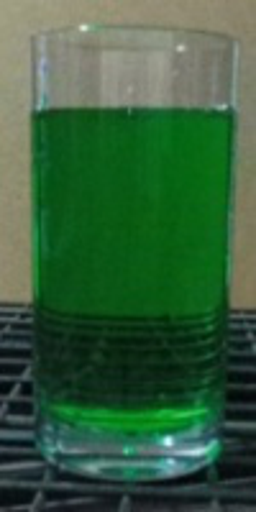}
    \end{subfigure}
    
    \par\smallskip 

    \begin{subfigure}[b]{0.15\columnwidth}
        \centering
        \includegraphics[width=\linewidth]{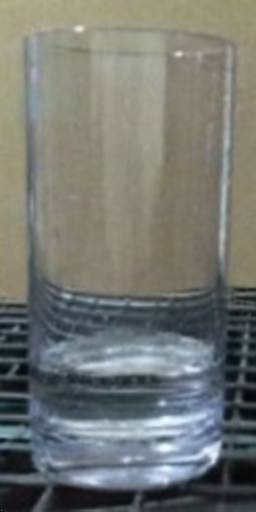}
    \end{subfigure}
    \begin{subfigure}[b]{0.15\columnwidth}
        \centering
        \includegraphics[width=\linewidth]{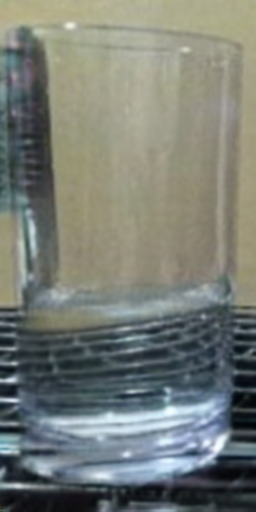}
    \end{subfigure}
    \begin{subfigure}[b]{0.15\columnwidth}
        \centering
        \includegraphics[width=\linewidth]{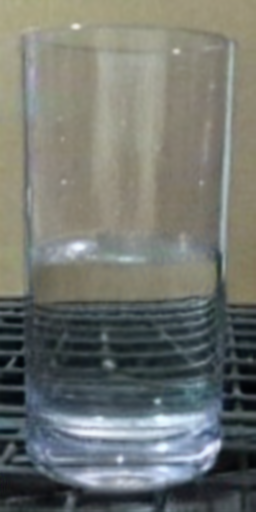}
    \end{subfigure}
    \begin{subfigure}[b]{0.15\columnwidth}
        \centering
        \includegraphics[width=\linewidth]{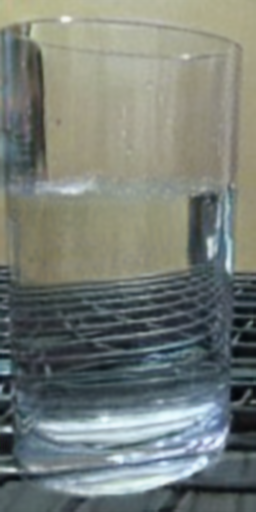}
    \end{subfigure}
    \begin{subfigure}[b]{0.15\columnwidth}
        \centering
        \includegraphics[width=\linewidth]{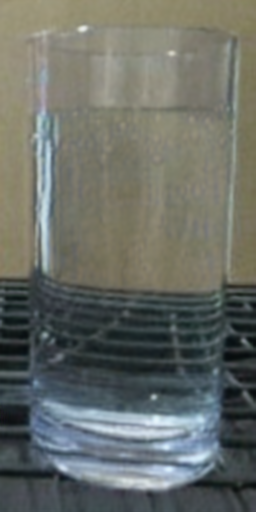}
    \end{subfigure}
    \begin{subfigure}[b]{0.15\columnwidth}
        \centering
        \includegraphics[width=\linewidth]{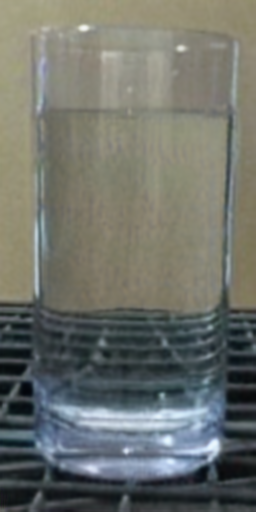}
    \end{subfigure}
    \caption{Image translation from colored liquid to transparent liquid by our trained model; Top Row: Real world colored liquid images; Bottom Row: Generated transparent liquid images.}
    \label{fig:visual_translation}
    
    \vspace{-1.5em}
\end{figure}

	We show representative examples of the image translation achieved by our CUT-based model in Figure~\ref{fig:visual_translation}. We observe that the network is able to translate the input image containing green-colored water pixels into images of clear water while still capturing the same background and refraction patterns as those in the source image. Most importantly, the transparent liquid in the generated images is in the same location as the colored liquid in the original images. This property allows us to apply the background mask from the colored liquids as the ground-truth label for the synthetic images of transparent liquids.

\subsubsection{Segmentation Performance}

To evaluate the segmentation performance of the method for transparent liquids, 
we compute the Intersection over Union (IoU) of predicted segmentations compared to the ground-truth. Our results for transparent liquid segmentation can be found in Table~\ref{tab:vision} (top row) as well as qualitatively in Figure~\ref{fig:segmentation_results} (bottom row).
Our method generally succeeds at segmenting the transparent liquid pixels in the image, achieving high IoU scores across the test set. Qualitatively, some of the predicted regions in Figure~\ref{fig:segmentation_results} are missing small patches on the interior, corresponding to turbulent or bubbly regions of the liquid that reflect the dynamic nature of the dataset. 

\begin{table}[ht]
    \centering
    \begin{tabular}{|c|c|c|c||c|}
        \hline
        \textbf{Method\textbackslash  IOU 	$\uparrow$} & \textbf{Low} & \textbf{Medium} & \textbf{High} & \textbf{All} \\ 
        \hline
        \hline
         Ours & 0.56 & 0.78 & 0.84 & 0.72  \\
         
        \hline
        \hline
         \multicolumn{5}{|c|}{\textit{Ablation 1}} \\ 
        \hline
         \hl{Color Jitter} & 0.02 & 0.04 & 0.06 & 0.03 \\ 
         \hline
         \hline
         \multicolumn{5}{|c|}{\textit{Ablation 2}} \\ 
         \hline
         Supervised (10\%) & 0.61 & 0.91 & 0.86 & 0.79  \\
         Supervised (1\%) & 0.52 & 0.54 & 0.38 & 0.50  \\
         Ours (10\%) & 0.56 & 0.80 & 0.78 & 0.71  \\
         Ours (1\%) & 0.38 & 0.60 & 0.53 & 0.51  \\

         \hline
    \end{tabular}
    \caption{Average Intersection over Union (IoU) scores on a test set of transparent liquid images, each filled with water to varying amounts.  We show the performance for subsets of images with varying amounts of liquid in the cup (Low, Medium, and High) as well as an average over all images. See Section \ref{sec:ablations} for descriptions of each of the ablations.
    }
    \label{tab:vision}
     \vspace{-1.5em}
\end{table}

\newcommand{\gridfig}[1]{\begin{subfigure}[t]{0.14\columnwidth}\includegraphics[width=\linewidth]{#1}\end{subfigure}}

\begin{figure}[t!]
    \centering
    
    \begin{minipage}[t]{0.2\columnwidth}
        \vspace{-3.5em}
        \textbf{Input}
    \end{minipage}
    \gridfig{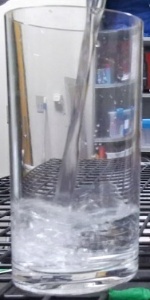}
    \gridfig{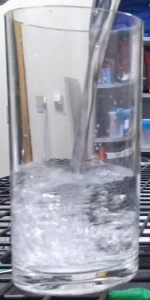}
    \gridfig{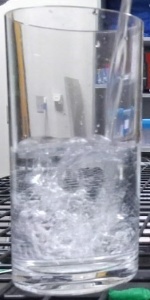}
    \gridfig{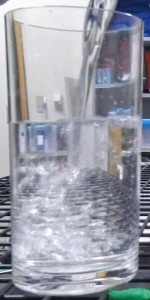}
    \gridfig{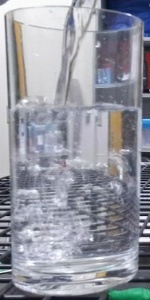}

    \par\smallskip
    
    \begin{minipage}[t]{0.20\columnwidth}
            \vspace{-4.5em}
        \textbf{Color \\Jitter}
    \end{minipage}
    \gridfig{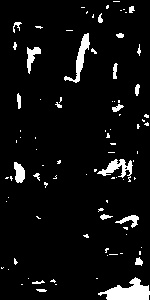}
    \gridfig{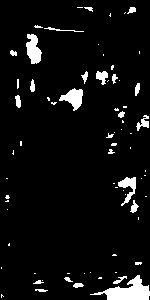}
    \gridfig{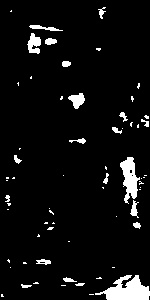}
    \gridfig{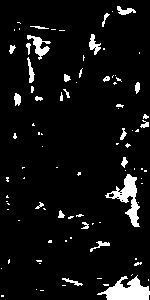}
    \gridfig{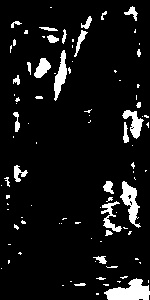}
    
    \par\smallskip
    
    \begin{minipage}[t]{0.20\columnwidth}
                \vspace{-4em}

        \textbf{Ours}
    \end{minipage}
    \gridfig{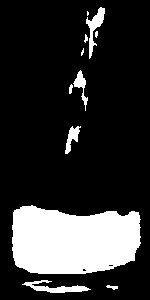}
    \gridfig{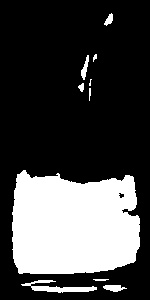}
    \gridfig{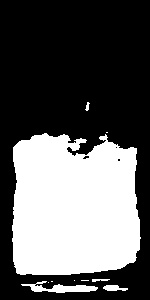}
    \gridfig{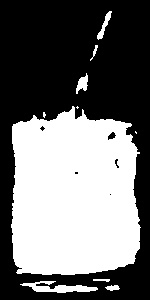}
    \gridfig{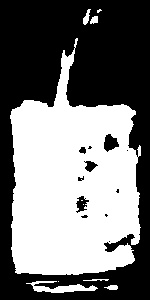}
    
    \caption{Representative segmentations of our method (bottom) compared to a model trained on images of colored liquid with color jitter augmentation \hl{(``Color Jitter'')}, which is unable to accurately segment the transparent liquid. }
    \label{fig:segmentation_results}
    \vspace{-1.5em}
\end{figure}

\subsubsection{Pouring Performance}

\label{sec:pouring_perf}
\begin{table}[ht]
    \centering
    \begin{tabular}{|c|c|c|}
        \hline 
        & \shortstack{Transparent Liquid \\ Ours (RMSE)} & \shortstack{Colored Liquid \\ Background Subtraction (RMSE)} \\
        
        \hline
        $0\% \rightarrow 25\%$ & $1.00 \pm 0.43\%$ & $8.46 \pm 2.14\%$ \\
        $0\% \rightarrow 50\%$ & $0.82 \pm 0.67\%$ & $1.86 \pm 0.71\%$ \\
        $0\% \rightarrow 75\%$ & $1.18 \pm 0.74\%$ & $1.57 \pm 0.50\%$ \\
        $25\% \rightarrow 75\%$ & $0.75 \pm 0.49\%$ & $1.61 \pm 0.65\%$ \\
        \hline
        All & $0.94 \pm 0.61\%$ & $3.38 \pm 3.18\%$ \\
        \hline
    \end{tabular}
    \caption{Percent error of the pouring system on both transparent and colored liquids. For transparent liquid pouring, we use our learned segmentation model; for colored liquid pouring, we use background subtraction. Contrary to expectations, our system performs better with transparent segmentation than with colored segmentation. }
    \label{tab:pouring}
     \vspace{-1em }
\end{table}


We conduct two sets of real-world robotic experiments to evaluate how our perception system performs on a real pouring task. We choose four different initial fill-levels and target fill-levels (see Table \ref{tab:pouring} ), and we conduct 20 pouring trials for each fill-level. We measure the level of liquid achieved upon the controller reaching the final \texttt{NotPouring} state, and we report the average error across each fill-level, as well as across all trials. We first evaluate with transparent water, using our transparent liquid segmentation system; and second with green-colored water, using the background-subtraction method described in the supplementary materials. Given the high accuracy of background subtraction, we feel this second task represents system performance with near-perfect segmentation. 

Results can be found in Table \ref{tab:pouring}.
In the case of pouring transparent liquids, we are able to achieve the desired ratio $l_{\mathrm{target}}$ with an average accuracy of 0.94\%, which corresponds to a roughly 0.13 cm error on average. Surprisingly, in the case of colored liquid (where background subtraction yields high-fidelity segmentation), pouring accuracy is worse, with an average error of 3.38\% or 0.47 cm. When observing the system, we noticed that the bounding box computation is sensitive to segmentations with splashing/sloshing: in these cases, the bounding box overestimates the amount of liquid in the cup and terminates pouring earlier than it should. Segmentations from our model pick up less of this transient liquid and thus outperform the segmentation obtained from background subtraction, when evaluated on the pouring task. 


\subsection{Visual Generalization}
\label{sec: generalization}


\newcommand{\ablationfig}[1]{\begin{subfigure}[t]{0.3\columnwidth}\includegraphics[width=\linewidth]{#1}\end{subfigure}}
\begin{figure}[ht!]
    \centering
    
    \begin{subfigure}[b]{.45\columnwidth}
            \includegraphics[width=\linewidth]{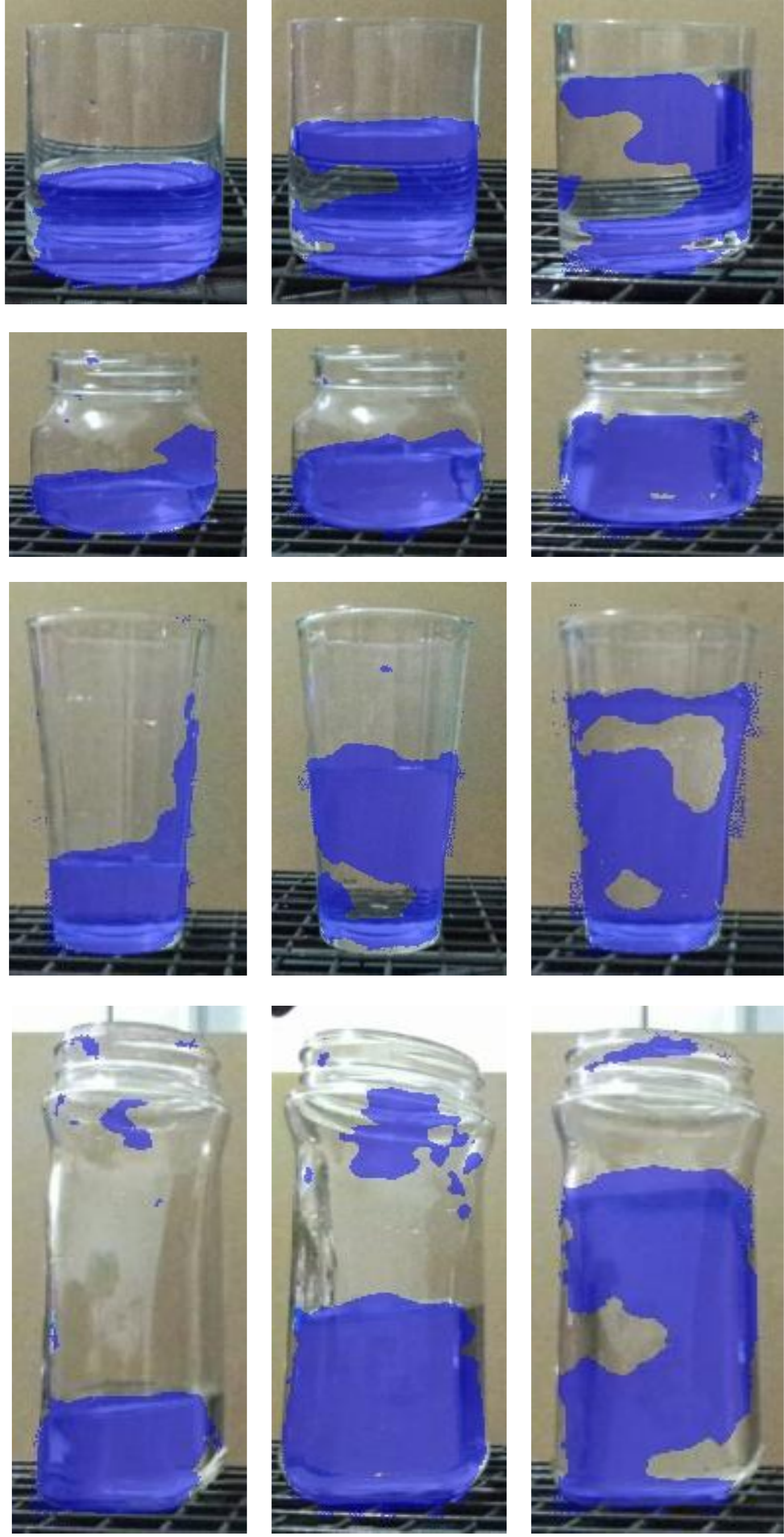}

        \caption{}
        \label{fig:novel_cups}

    \end{subfigure}
    \hfill
    \begin{subfigure}[b]{.45\columnwidth}
          \includegraphics[width=\linewidth]{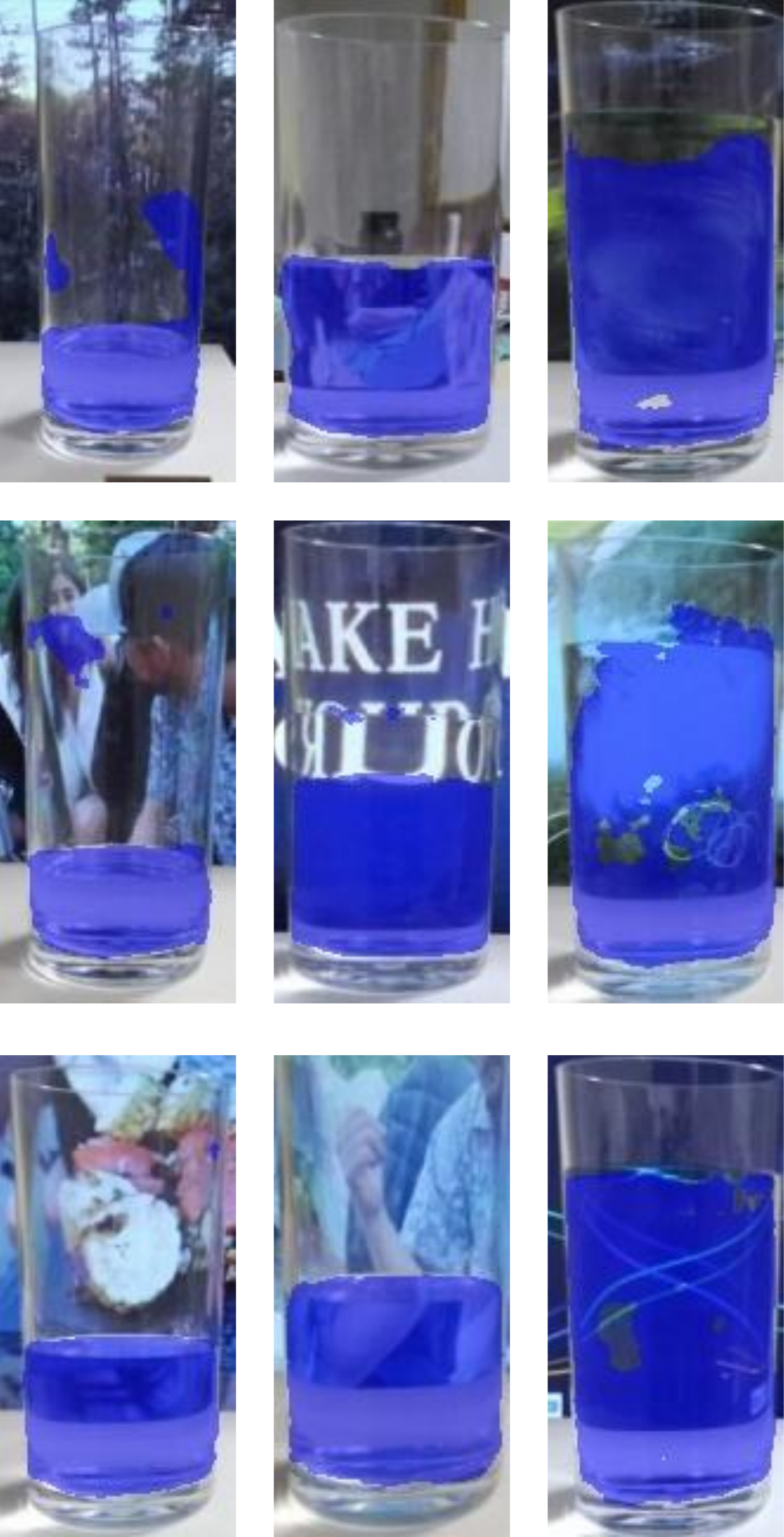}

          

          

          

           

          
        \caption{}
          \label{fig:novel_backgrounds}

    \end{subfigure}

    \caption{\hl{Generalization of our transparent liquid segmentation model to (a) unseen cups, and (b) unseen backgrounds.}}
    \label{fig:generalization}
\end{figure}

In this section, we explore the potential of our method to segment transparent liquids in more diverse settings. 

\subsubsection{Diverse containers} 
We first train a segmentation model on an image dataset consisting of a single cup in a single scene with varying heights of liquids. We then evaluate on 7 transparent cups (73 images) with different shapes and different fill levels of transparent liquid. As can be seen in Figure \ref{fig:novel_cups} \hl{and in Table}~\ref{tab:generalization}, the model generalizes reasonably to other cups, indicating that it has learned to detect relevant liquid features invariant to a specific container.

\subsubsection{Diverse backgrounds}

 Next, we train a model on a cup in front of a diverse set of backgrounds and various water heights. Because transparent containers and liquid will refract the background, we cannot easily imitate this effect synthetically using simple approaches such as pasting a foreground segment on top of a random background scene \cite{kisantal2019augmentation,dvornik2018modeling,dvornik2019importance}. Instead, to generate a diverse set images with physically-accurate refraction characteristics, we set up a large flatscreen LED display behind the pouring scene and play videos containing diverse indoor and outdoor scenes during data collection. This allowed us to create datasets with a high degree of background diversity, with natural patterns of light refracting through water. We evaluate our method on 90 images having cups filled to various heights placed in front of \hl{unseen videos}. As can be seen in Figure \ref{fig:novel_backgrounds} \hl{and in Table}~\ref{tab:generalization}, the model learns a segmentation function that is invariant to  unseen backgrounds.
 These two experiments demonstrate the potential of our approach to scale to a general set of scenes.

\begin{table}[ht]
    \centering

\begin{tabular}{|c|c|c|c||c|}
        \hline
        \textbf{Variation} & \textbf{Low} & \textbf{Medium} & \textbf{High} & \textbf{All} \\ 
        \hline 
     \hl{Diverse Containers} & 0.67 & 0.68 & 0.60 & 0.65\\
     \hl{Diverse Backgrounds} & 0.70 & 0.81 & 0.77 & 0.76\\
     \hline
\end{tabular}
    \caption{
    \hl{Average IoU scores on test sets of transparent liquid images placed in a) novel containers of varying shaped and size and b) unseen backgrounds. See Section} \ref{sec: generalization} \hl{for more details}.}
    \label{tab:generalization}
    \vspace{-1.5em}
\end{table}


\subsection{Visual Ablations}
\label{sec:ablations}

\noindent \textbf{Ablation 1: What is the benefit of training the segmentation model on synthetically generated transparent (instead of colored) liquid?} To explore whether data augmentation on colored liquids is sufficient to train a model for transparent liquid segmentation, we train a segmentation model on colored liquid with color jitter and evaluate it on transparent liquid. We jitter the brightness, contrast and hue to obtain the input image for training the segmentation model.
Results  are shown in Table~\ref{tab:vision} \hl{(``Color Jitter'')}  and  Fig~\ref{fig:segmentation_results}. The model fails to detect the correct liquid height during our evaluation, showing that using color jitter on a colored liquid image is not sufficient domain randomization to capture the texture and patterns required to segment transparent liquids.

\par\smallskip

\noindent \textbf{\hl{Ablation 2: What is the benefit of self-supervised learning on a large unlabeled dataset over training on a small  manually-labeled dataset?}} To evaluate the benefit of training on a larger dataset of \hl{synthetically generated} transparent liquid images (generated using our image translation model), we compare with models trained on smaller datasets of manually-annotated images of transparent liquid, to approximate what a researcher might be able to accomplish with a limited amount of annotation time. We manually annotate two subsets of our transparent liquid dataset: one containing 220 images (10\%), and one containing 22 images (1\%). To see how well our method trained on synthetic images performs at similar sample sizes, we additionally split our synthetic dataset into 10\% and 1\% subsets. We train a segmentation model on each. Results are found in Table \ref{tab:vision}.

We find that performance on the segmentation task saturates when roughly 10\% of the available data is used: ``Ours (10\%)'' shows the same performance as our full method (``Ours''), and ``Supervised (10\%)'' outperforms our method. However, with only 1\% of manually-annotated labels available, the supervised baseline (``Supervised (1\%)'') performs substantially worse than our full method. \hl{As we move towards scaling to more diverse scenes, it will become increasingly difficult to manually annotate all of the data that is needed; in such a case, our self-supervised method can enable scalable training without manual annotations.}


\par\smallskip



\section{CONCLUSION}

In this paper, we propose a method to segment transparent liquid placed inside transparent containers using static RGB images. A generative model is used to translate colored liquids to transparent liquids, without any manual annotation or alignment. We show that a network trained on these \hl{generated} images can be used to segment \hl{real} transparent liquids directly from RGB images.  Finally, we demonstrate the utility of transparent liquid segmentation on a real robotic pouring task. In future work, we would like to scale up our transparent liquid perception model to a wider variety of settings, with \hl{a single model trained over} diverse containers, backgrounds, lighting conditions, and camera angles. We hope that our method paves the way for more flexible and robust perception of transparent liquids for robot pouring. 
\section*{ACKNOWLEDGMENT}
This material is based upon work supported by LG Electronics and National Science Foundation under Grant No. IIS2046491.



\bibliographystyle{IEEEtran}
\bibliography{IEEEabrv,references}





\renewcommand\hl[1]{#1} 

\newcommand{\figurename}{Supplementary Figure}

\appendix

\begin{figure*}[h]
    \centering
    \begin{tabular}{m{0.15\textwidth} m{0.15\textwidth} m{0.15\textwidth} m{0.15\textwidth} m{0.15\textwidth}}
        \includegraphics[width=0.15\textwidth]{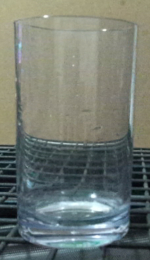} &
        \includegraphics[width=0.15\textwidth]{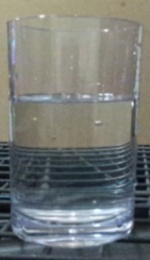} &
        \includegraphics[width=0.15\textwidth]{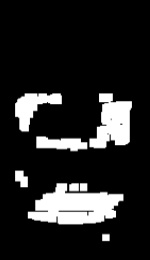} &
        \includegraphics[width=0.15\textwidth]{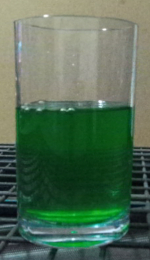} &
        \includegraphics[width=0.15\textwidth]{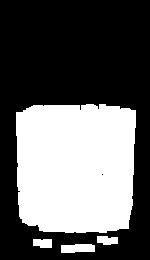} 
        \\
        \textbf{(a)} &
        \textbf{(b)} &
        \textbf{(c)} &
        \textbf{(d)} &
        \textbf{(e)} 
 \end{tabular}
    \caption{Background subtraction for automatic ground truth annotation of colored liquid: (a) Empty cup from which we build the background model (b) Cup filled with transparent liquid (c) Background subtraction for transparent liquid; note that this mask is not accurate and hence our method does not rely on background subtraction of transparent liquid (d) Cup filled with colored liquid (e) Background subtraction for colored liquid; this mask is much more accurate than the one in (c), hence we use this mask to train the transparent liquid segmentation model, trained on synthetically generated images of transparent liquid.} 
    \label{fig:bgsb_visualization}
\end{figure*}

\subsection{Background Subtraction}

Obtaining liquid vs non-liquid labels from RGB images usually requires human annotations; however, such annotations are tedious to obtain. Instead, we propose to automatically generate ground truth annotations for segmentation using background subtraction.

To do so, we record a set of images of an empty cup (Fig~\ref{fig:bgsb_visualization}a); on these images, we train a background subtraction model.  Specifically, we use a Gaussian Mixture Model (GMM) that automatically estimates the number of components for each pixel~\cite{bgsb_model_opencv}. Then, without moving the cup, we fill the cup with colored liquid (Fig~\ref{fig:bgsb_visualization}d). We apply the background subtraction model to estimate the visual difference between the empty and full cup (of colored liquid). This difference provides a ground-truth segmentation label (Fig~ \ref{fig:bgsb_visualization}e) for the location of the colored liquid in the cup. We will refer to this segmentation label as $I_{gt}$. We will later use this segmentation mask to learn to segment transparent liquid.

If we wish to segment transparent liquids, one might ask that why we don't apply background subtraction directly to the images of transparent liquid? As observed in Fig~\ref{fig:bgsb_visualization}c, background subtraction is not able to detect the transparent liquid pixels between Fig~\ref{fig:bgsb_visualization}a \& Fig~\ref{fig:bgsb_visualization}b. This is because the GMM detects if the difference from the background model exceeds a threshold. Transparent liquids do not offer sufficient visual difference in the image to exceed this threshold; reducing the threshold too low will lead to noisy foreground estimation. Instead, we perform background subtraction only on images of colored liquids, which is relatively easy and accurate, as shown in Fig~\ref{fig:bgsb_visualization}d and e.

\subsection{Transparent Liquid Segmentation}

We train a transparent liquid segmentation model from scratch to classify liquid pixels in an image. We use paired samples of synthetic transparent liquid images and ground truth obtained from background subtraction to train the segmentation model. The segmentation model is trained using the Binary Cross Entropy loss between the predicted liquid segmentation mask and the ground truth. We use a UNet style architecture with four downsampling blocks followed by four upsampling blocks \hl{and has about 13 million trainable parameters.} We use stochastic gradient descent to train the network with learning rate 1e-3, momentum 0.9 and weight decay 5e-4.\hl{ It takes us ~5 hours to train both the image translation and segmentation model sequentially.}

The input to each downsampling block is passed through a 2D MaxPool operator followed by two convolutional layers. The output from each convolutional layer is batch normalized and passed through a ReLU activation function. Each upsampling block takes a concatenated input of bilinear interpolation from the previous upsampling layer and a skip connection from the downsampling layer.

\subsection{Vision System}

For our cup detection method we use Translab~\cite{translab} with scale width option to keep the aspect ratio same for all input images. We detect the target cup location at the beginning of the pour and do not change it's position during and after the pour.

We crop the image around the target cup using OpenCV's minAreaRect function along with a padding of 10 pixels to allow some background information to the network. The cropped image is segmented using the UNet network at 10 Hz on an Nvidia 2080 Ti GPU. \hl{We use OpenCV's morphological expansion filter (erosion + dilation) with an identity kernel of size 5 to remove noise from the resulting segmentation mask. In order to pour to a target height we require information about the volume of water that has already reached the cup and not water present in the pouring stream, for this we use OpenCV to find the largest contour in the liquid segmentation mask and retain water pixels within this region. We usually start the liquid segmentation method ~2 seconds after pouring has started to ensure that there is some water in the target cup.}

\hl{The post-processed} liquid segmentation mask is then used to detect water height using OpenCV's minAreaRect function. The top edge of the detected bounding box becomes the predicted water height (process variable) for our control algorithm.

\subsection{Generalization}

\hl{In order to evaluate our method's capacity to generalize to diverse situations lying outside the training distribution, we design experiments to test the the trained segmentation models on scenes with 1) \textit{various} transparent containers filled with water to different heights, and 2) same transparent container filled with water to different heights in front of different backgrounds.}

\subsubsection{Diverse containers}

\begin{figure*}[h]
    \centering
    \includegraphics[width=0.8\textwidth]{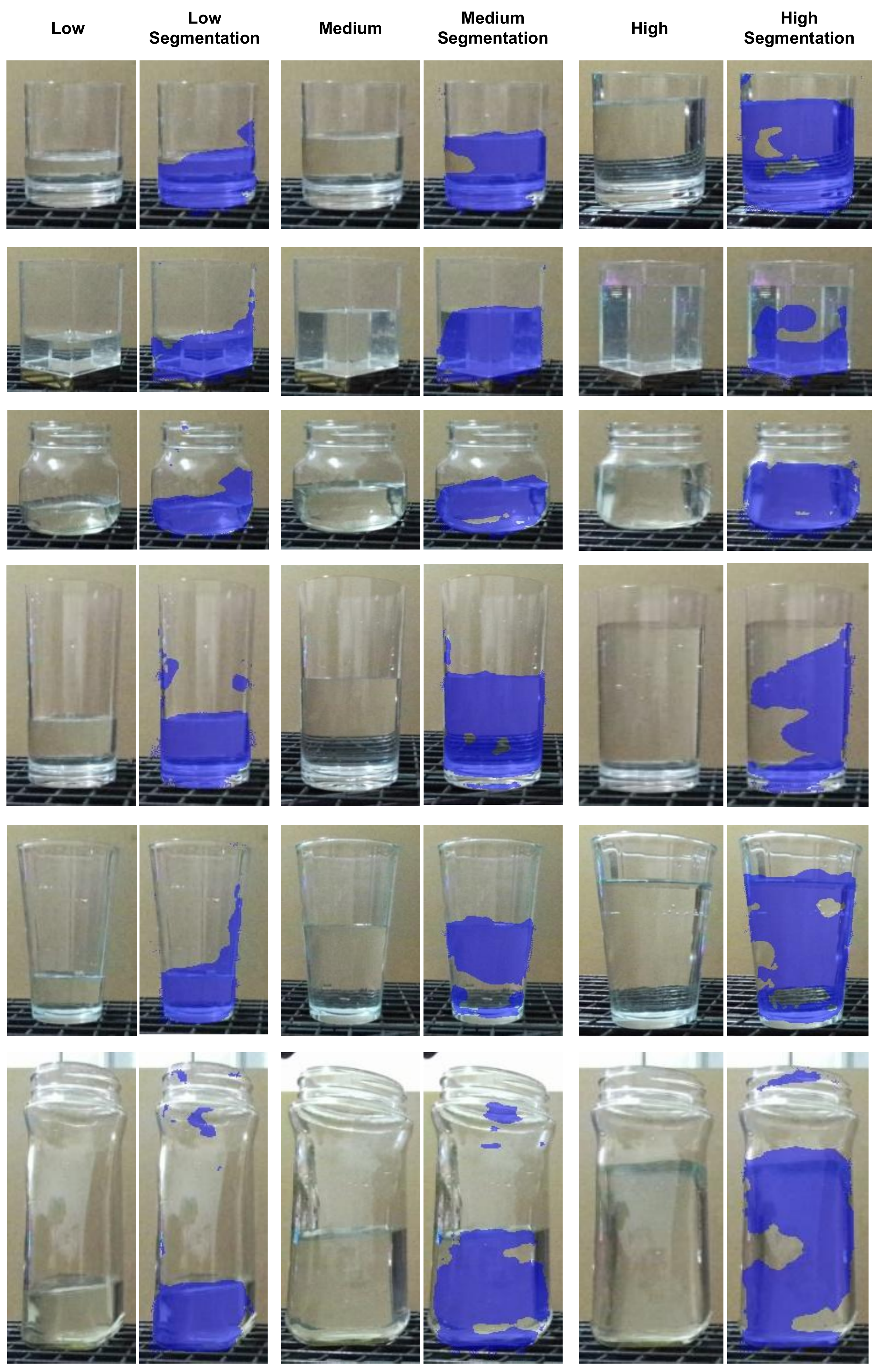}
    \caption{We train our method on one container (circular cup) and evaluate on unseen containers having various shapes, sizes and liquid quantities. Our results show generalization of our transparent liquid segmentation method to cups and jars that have completely different shapes (wide, tall, square, conical, oval and with spout) from the cup used during training}
    \label{fig:diverse_cup_shapes}
\end{figure*}

\hl{We train our method using 128 images of a single cylindrical cup filled with water to different heights and test on 73 images of 7 different containers filled to different heights of water. The containers used for testing have different shapes, sizes and spouts; this causes them to produce various refraction patterns. We also test with cups placed in different locations of the scene to verify that the model does not overfit to a particular cup pose.

We observe that the model is able to segment clear water within the transparent container in most cases as shown in Fig}~\ref{fig:diverse_cup_shapes}.

\subsubsection{Diverse Backgrounds}
\begin{figure*}[h]
    \centering
    \includegraphics[width=0.8\textwidth]{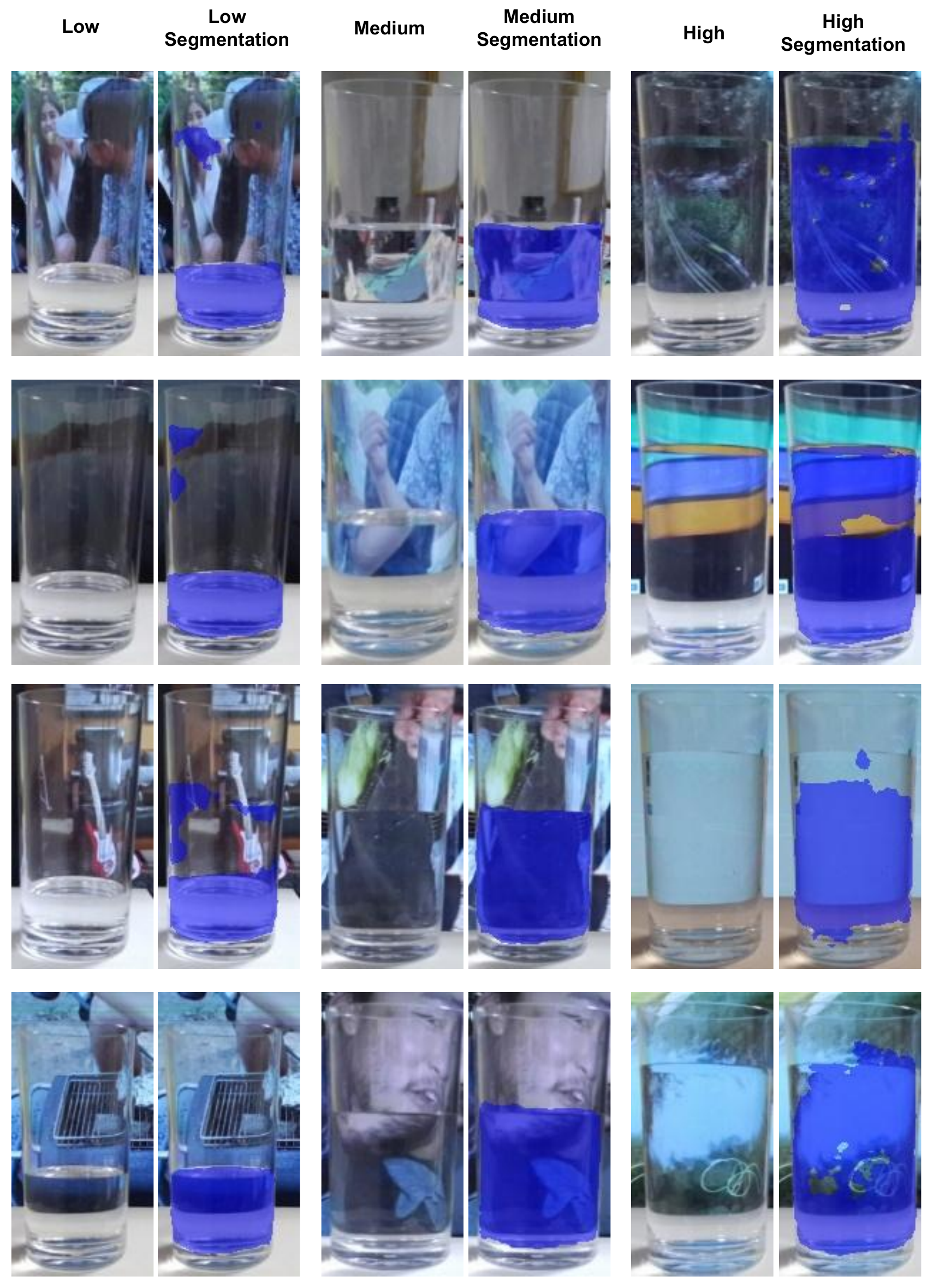}
    \caption{We show evaluations for transparent liquid segmentation when liquid is placed in front of various backgrounds. We show segmentation masks (blue) for short, medium and tall test cases based on the quantity of liquid placed in the container.}
    \label{fig:diverse_backgrounds}
\end{figure*}

\hl{
We collect two datasets of 2000 images (sampled from 20 vidoes) each of transparent and colored liquid placed in front of diverse backgrounds. We are able to generate diverse backgrounds by placing a computer monitor playing random youtube videos behind the cup. We then manually pour water and record images with different amounts of static water in the cup. We follow the same pipeline as described in Section III of the main text., i.e train an image translation model to translate colored liquid images to transparent liquid and then use the translated images for segmentation.

Our evaluation procedure utilizes 500 images sampled from 5 random youtube vidoes. Segmentation results can be found in Fig}~\ref{fig:diverse_backgrounds}.

\end{document}